\documentclass[11pt,a4paper]{article}
\usepackage[hyperref]{eacl2021}
\usepackage{times}
\usepackage{latexsym}
\usepackage{graphicx}
\usepackage{tabularx}
\usepackage{soul}
\usepackage{hyperref}
\usepackage{url}
\usepackage{booktabs}

\usepackage{microtype}

\aclfinalcopy 
\def\aclpaperid{1259} 

\title{Cognition-aware Cognate Detection}

\author{Diptesh Kanojia\textsuperscript{$\dagger$,$\clubsuit$,$\star$}, Prashant Sharma\textsuperscript{$\diamond$}, Sayali Ghodekar\textsuperscript{$\ddagger$},\\ \textbf{Pushpak Bhattacharyya}\textsuperscript{$\dagger$},
\textbf{Gholamreza Haffari}\textsuperscript{$\star$}, and \textbf{Malhar Kulkarni}\textsuperscript{$\dagger$}\\
  \textsuperscript{$\dagger$}IIT Bombay, India,  \textsuperscript{$\diamond$}Hitachi CRL, Japan, \textsuperscript{$\ddagger$}RingCentral, India \\
  \textsuperscript{$\clubsuit$}IITB-Monash Research Academy, India \\
  \textsuperscript{$\star$}Monash University, Australia \\
  {\tt \textsuperscript{$\dagger$}\{diptesh,pb,malhar\}@iitb.ac.in, \textsuperscript{$\diamond$}prashaantsharmaa@gmail.com}\\
  {\tt \textsuperscript{$\ddagger$}sayalighodekar26@gmail.com, \textsuperscript{$\star$}gholamreza.haffari@monash.edu} \\}

\date{25-01-2021}

\begin{document}
\maketitle
\begin{abstract}

Automatic detection of cognates helps downstream NLP tasks of Machine Translation, Cross-lingual Information Retrieval, Computational Phylogenetics and Cross-lingual Named Entity Recognition. Previous approaches for the task of cognate detection use orthographic, phonetic and semantic similarity based features sets. In this paper, we propose a novel method for enriching the feature sets, with cognitive features extracted from human readers' gaze behaviour. We collect gaze behaviour data for a small sample of cognates and show that extracted cognitive features help the task of cognate detection. However, gaze data collection and annotation is a costly task. We use the collected gaze behaviour data to predict cognitive features for a larger sample and show that predicted cognitive features, also, significantly improve the task performance. We report improvements of 10\% with the collected gaze features, and 12\% using the predicted gaze features, over the previously proposed approaches. Furthermore, we release the collected gaze behaviour data along with our code and cross-lingual models. 

\end{abstract}

\section{Introduction}

Cognates are word pairs, across languages, which have a common etymological origin. For example, the French and English word pair, \textit{Libert{\'e} - Liberty}, reveals itself to be a cognate through orthographic similarity. The task of automatic cognate detection across languages\footnote{Cognates can also exist in the same language. Such word pairs/sets are commonly referred to as \textit{doublets}.} requires one to detect word pairs which are etymologically related, and carry the same meaning. Such word pairs share a formal and/or semantic affinity and facilitate the second language acquisition process, particularly between related languages. Although they can accelerate vocabulary acquisition, language learners also need to be aware of false friends and partial cognates, at times, leading to unrelated semantic coupling. For example, \textit{``gift''} in German means \textit{``poison''}, which is a known example of a False Friend pair. For an example of a partial cognate, the word \textit{``police''} can translate to \textit{``police''}, \textit{``policy''} or \textit{``font''}, in French, depending on the context in which it was used.

Manual detection of such cognate sets requires a human expert with a good linguistic background in multiple languages. Moreover, manual annotation of cognate sets is a costly task in terms of time and human effort. Automatic Cognate Detection (ACD) is a well-known task, which has been explored for a range of languages using different methods; and has shown to help Cross-lingual Information Retrieval~\cite{meng2001generating}, Machine Translation (MT)~\cite{al1999statistical}, and Computational Phylogenetics~\cite{rama2018automatic}. In the traditional approaches to automatic cognate detection, words with similar meanings or forms are used as probable cognates~\cite{jager2017using,rama2016siamese,kondrak-2001-identifying}. From such pairs, the ones that exhibit a high phonological, lexical and/or semantic similarity, are analyzed in order to find true cognates.~\citet{merlo-andueza-rodriguez-2019-cross} perform an investigation to evaluate the use of cross-lingual embeddings and show that these models inherit a lexical structure matching the bilingual lexicon. This study establishes that cross-lingual models can provide an effective similarity score compared to their monolingual counterparts, for both cognates and false friends. However, they do not evaluate machine learning (ML) approaches that distinguish between cognates and false friends. The absence of such an evaluation motivates us to use cross-lingual similarity scores with ML algorithms. Our work\footnote{\href{http://www.cfilt.iitb.ac.in/eacl2021diptesh}{Dataset and Code}} reports whether these scores can provide an adequate distinction or not.

Inspired from their work, we investigate the use of cross-lingual word embeddings and cognitive features to distinguish between cognates and false friends, for the Indian language pair of Hindi and Marathi. Cognitively inspired features have shown to improve various NLP tasks~\cite{mishra2018cognition}. However, most of their work involves collecting the gaze behaviour data first on a large sample, and then splitting the data into training and testing data, before performing their experiments. While their work does show significant improvements over baseline approaches, across multiple NLP tasks, collecting gaze behaviour over a large cognate dataset can be costly, both in terms of time and money. Our approach tries to reduce annotation cost by predicting gaze behaviour data for a large sample based on the smaller sample of collected gaze data. Our investigations use three recently proposed cross-lingual word embeddings based approaches to generate features for the task of cognate detection. We also generate cognitive features from participants' gaze behaviour data and show that gaze behaviour helps the task of cognate detection. Additionally, we use the collected gaze behaviour data and predict gaze-based features for a much larger sample. We believe that using gaze features will be more applicable only if gaze features can be predicted for unseen samples. We believe that collection of gaze data cannot be performed in all the scenarios and hence hypothesize that predicting such data if it helps improve the task of cognate detection, should be a viable solution.

\begin{table}[]
\centering
\resizebox{\columnwidth}{!}{%
\begin{tabular}{@{}ccccc@{}}
\toprule
 & \textbf{Hindi (Hi)} & \textbf{Marathi (Mr)} & \textbf{Hindi Meaning} & \textbf{Marathi Meaning} \\ \midrule
\textbf{Cognate} & ank & ank & Number & Number \\
\textbf{False Friend} & shikshA & shikshA & Education & Punishment \\ \bottomrule
\end{tabular}%
}
\caption{An example each of a cognate and a false friend pair from Indian languages, Hindi and Marathi.}
\vspace{-1em}
\label{tab:examples}
\end{table}

\subsection*{Motivation}
Consider a scenario where an NLP task comes across the false friend pair in Table \ref{tab:examples}. Orthographic similarity or even phonetic similarity-based techniques will fail to detect the difference between the Hindi meaning of the word ``shikhshA'' and its Marathi counterpart. Here, semantic approaches should detect the distinction in meaning, but monolingual embeddings are trained using a large corpus \textbf{from the same language}. In such cases, it becomes imperative that a cross-lingual word embeddings model be utilized. However, Indian languages are known to be low-resource languages compared to English or even many European languages like French, Italian, German \textit{etc.} Acquiring additional clean data for training cross-lingual models is, yet again, a painful task. In such a scenario then, we ask ourselves, \\
\\
\textit{``Can cognitive features be used to help the task of Cognate Detection?''} \\ 
\\
furthermore, \\
\\
\textit{``Using gaze features collected on a small set of data points, can we predict the same features on a larger set of data points to alleviate the need for collecting gaze data?''}\\

With this work, we try to answer both the questions stated above and present the rest of the paper as follows. We discuss the current literature on the cognate detection task and cognitive NLP in Section 2. Section 3 discusses the dataset acquisition, including the description and analysis of our gaze behaviour data. We describe the feature sets used in section 4. Our approaches to the task of cognate detection are discussed in Section 5. The results of our work are discussed in Section 6. Finally, we conclude the study with possible future work in this direction in Section 7.

\subsection*{Terminology}

An \textit{interest area (IA)} is an area of the annotation screen to be processed by the human reader. In our experiments, it is an area where a word is shown to the reader. A \textit{fixation} is an event where the reader focuses within an ``interest area''. A \textit{saccade} is the movement of the eye from one fixation point to another. If the saccades move from an earlier IA to a later IA, such a saccade is called a \textit{progression}.  A \textit{regression} is the saccade path when the reader moves back to a previous IA. We also use the terms \textit{reader} and \textit{participant} interchangeably. 

\section{Related Work}

Current literature which uses gaze behaviour to solve downstream NLP tasks has been applied to the NLP tasks of sentiment analysis~\cite{mishra2018cognition,barrett-etal-2018-sequence,long2019improving}, sarcasm detection~\cite{mishra-etal-2016-harnessing}, grammatical error detection~\cite{barrett-etal-2018-sequence}, hate speech detection~\cite{barrett-etal-2018-sequence}, named entity recognition~\cite{hollenstein-zhang-2019-entity}, part-of-speech tagging~\cite{barrett-etal-2016-pos-tagging}, sentence simplification~\cite{klerke-etal-2016-improving}, and readability~\cite{gonzalez2018learning,singh-etal-2016-quantifying}. A comprehensive overview is provided by~\newcite{mishra2018cognitively}. The primary motivation of using cognitive features for NLP tasks is derived from the eye-mind hypothesis~\cite{just1980theory}, which establishes a direct correlation between a reader's comprehension of the text with the time taken to read the text. This hypothesis has initiated a large body of psycholinguistic research that shows a relationship between text processing and gaze behaviour.  \citet{yaneva2020classifying} discuss the use of gaze features for the task of anaphora resolution. Their findings show that gaze data can substitute the classical text processing approaches along with the fact that human disambiguation process overlaps with the information carried in linguistic features. \citet{rohanian-2017-multi} use gaze data to automatically identify multiword expressions and observe that gaze features help improve the accuracy of the task when combined with traditional linguistic features used for the task.

~\citet{mathias2020happy} describe an approach to scoring essays in a multi-task learning framework automatically. Their approach relies on collecting gaze behaviour for essay reading for a small set of essays and then predicting the rest of the dataset's gaze behaviour in a multi-task learning setup. Similarly,~\citet{barrett-etal-2016-pos-tagging} use token level averages of cognitive features at run time, to mitigate the need for these features at run time.~\citet{singh-etal-2016-quantifying} and~\citet{long2019improving} predict gaze behaviour at the token-level as well.~\citet{mishra2018cognition},~\citet{gonzalez2018learning},~\citet{barrett-etal-2018-sequence}, and~\citet{klerke-etal-2016-improving}, use multi-task learning to solve the primary NLP task, where learning the gaze behaviour is an auxiliary task.

Orthographic/String similarity-based methods are often used as baseline methods for the cognate detection task, and the most commonly used method amongst them is the Edit distance-based similarity measure~\cite{melamed1999bitext,mulloni2006automatic}. Research in automatic cognate detection using various aspects involves the computation of similarity by decomposing phonetically transcribed words~\cite{kondrak2000new,dellert2018combining}, acoustic models~\cite{mielke2012assessing}, clustering based on semantic equivalence~\cite{hauer2011clustering}, and aligned segments of transcribed phonemes~\cite{list2012lexstat}.~\citet{rama2016siamese} employs a Siamese convolutional neural network to learn the phonetic features jointly with language relatedness for cognate identification.~\citet{jager2017using} use SVM for phonetic alignment and perform cognate detection for various language families. Various works on orthographic cognate detection usually take alignment of substrings within classifiers like SVM~\cite{ciobanu2014automatic,ciobanu2015automatic} or HMM~\cite{bhargava2009multiple}.~\citet{ciobanu2014automatic} employ dynamic programming based methods for sequence alignment. Cognate facilitation in second language learners has previously been explored with the help of eye-tracking studies~\cite{blumenfeld2005covert,van2009does,bosma2020cognate} but the task of cognate detection has not been performed with gaze features obtained from the cognitive data in any of the previously available literature. 

For the task of cognate detection, however, the use of cognitive features has not been established previously. The task of cognate detection is cross-lingual, and a reader's cognitive load should vary while trying to comprehend the meaning of concepts, from different languages. Our work tries to exploit the difference noted in terms of time taken and eye-movement patterns in cognates \textit{vs} false friends, to generate additional features for the ACD task. Moreover, for Indian languages such as Marathi, where agglutination\footnote{Agglutination is a linguistic process pertaining to derivational morphology in which complex words are formed by stringing together morphemes without changing them in spelling or phonetics.} varies the word length, the task becomes tougher, computationally.

\section{Dataset Acquisition \& Analysis}
\label{sec:dses}

We pose the problem of cognate detection as a binary classification task in a supervised setting. We use a recently released challenging dataset~\cite{kanojia2020challenge} of cognates and false friends. We extract the Hindi-Marathi cognate and false friend pairs. The number of cognate and false friend pairs released by the paper cited above is 15726, and 5826. We select an equal number of cognates at random to reduce this skew, thus producing a balanced dataset for the classification task. For any further experiments in our paper, we use this artificially class-balanced dataset of 5826 (cognates) + 5826 (false friends) data points. We also augment the complete dataset with context information from the IndoWordnet~\cite{bhattacharyya2017indowordnet}. The context information contains a \textit{gloss} and an \textit{example sentence} from the Wordnet data. The dataset released by~\citet{kanojia2020challenge} contains the Synset IDs for each word pair, which helps us locate exact concept information from the dataset. We provide positive labels to cognates and negative labels to false friend pairs obtained from this data and construct what we call ``D1''. 

\begin{figure*}[ht!]
    \centering
    \includegraphics[width=1\textwidth]{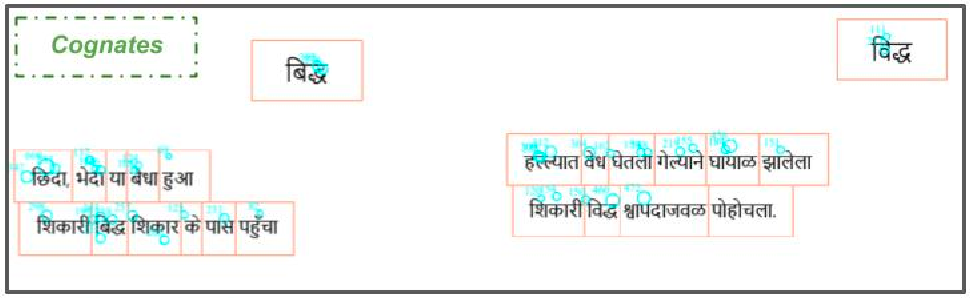}
    \caption{Screen capture showing collection of gaze features (via eye tracking) while displaying word pairs along with respective definitions and examples. The figure shows the cognate pair (\textit{BiDha} - \textit{BiDha}) where both mean ``pierced'' in the context of hunting. The figure also shown the glosses and example sentences provided to the annotator for this cognate pair in their respective languages (Hindi and Marathi)}
    \label{fig:eyetracking}
\end{figure*}

We extract 100 pairs, at random, from each of the positive and negative labels for collecting gaze behaviour data, to construct what we call ``D2''. This data, extracted from D1, is used to collect gaze behaviour and annotation. Although we have gold labels for the data extracted, we ask the participants to annotate the data by asking them if the concepts shown on the screen mean the same. The annotation screen provides them with contextual clues obtained from Wordnet data on the screen, as shown in Figure \ref{fig:eyetracking}. The complete dataset statistics are shown in Table \ref{tab:data}.

\begin{table}[]
\centering
\resizebox{\columnwidth}{!}{%
\begin{tabular}{@{}ccc@{}}
\toprule
 & \textbf{Cognates (1)} & \textbf{False Friends (0)} \\ \midrule
Kanojia et. al. (2020) & 15726 & 5826 \\
\textbf{D1} & \textbf{5826} & \textbf{5826} \\
\textbf{D2} & \textbf{100} & \textbf{100} \\ \bottomrule
\end{tabular}%
}
\caption{Dataset Statistics for Cognate Detection Task}
\vspace{-1em}
\label{tab:data}
\end{table}

\subsection{Gaze Data Collection and Annotation}

The task assigned to annotators was to read word pairs and the contextual clues provided on the screen, one pair at a time. The annotators were requested to label the pairs with a binary score indicating the similarity in meaning (\textit{i.e.,} positive/negative). It should be noted that the participants were not instructed to annotate whether a pair is a cognate or a false friend, to rule out the Priming Effect, (\textit{i.e.,} if the exact task (cognates \textit{vs} false friends) is expected beforehand, processing cognate pairs will become relatively easier~\cite{sachez1992bilingual}). This ensures the ecological validity of our experiment in two ways: (1) The participant does not have any clue so that they can treat cognates with special attention (done by asking them to annotate based solely on meaning similarity) (2) Cognate pairs are mixed with false friend pairs and the participant does not have any prior knowledge about whether the next word pair would be a cognate or not. This also ensures that the participants pay attention to the task and do not just skim through the word-pair presented on the screen.

It should be noted that all our participants are primarily Marathi speakers and have learnt Hindi at the school level. Hindi and Marathi are considered to be relatively closer languages due to their shared vocabulary and the geographical location of the demographic. All the participants speak, understand and write - both Hindi and Marathi.

The collection of gaze data is conducted by following the standard norms in eye-movement research~\cite{holmqvist2011eye}. While reading, an SR-Research Eyelink-1000 eye-tracker (monocular remote mode, sampling rate 500Hz) records several eye-movement parameters like fixations (a long stay of gaze) and saccade (quick jumping of gaze between two positions of rest) and pupil size. For this experiment, the default value of 4ms was used for a gaze to be counted as fixation. We request a total of 15 participants to perform the annotation task, out of which only 11 participants could perform the data collection\footnote{We could not perform gaze data collection with the remaining 4 participants due to the COVID-19 pandemic.}.

Out of the 11 completed annotations, we discarded the data from 2 participants as their gaze behaviour was erratic (the fixations were too far away from the IAs). The participants are graduates with science and engineering background. They are bilingual speakers who know both Hindi and Marathi. Our participants were given a set of instructions beforehand and were advised to seek clarifications before they proceed. The instructions mention the nature of the task as discussed above, annotation input method, and the necessity of head movement minimization during the experiment.

\begin{table}[]
\centering
\resizebox{\columnwidth}{!}{%
\begin{tabular}{@{}cccccc@{}}
\toprule
\multicolumn{1}{l}{} & \textbf{$\mu$\_Pos} & \textbf{$\sigma$\_Pos} & \textbf{$\mu$\_Neg} & \textbf{$\sigma$\_Neg} & p \\ \midrule
P1 & 9.720 & 17.867 & 8.677 & 4.281 & 0.028 \\
P2 & 8.596 & 10.526 & 7.619 & 13.794 & 0.049 \\
P3 & 7.770 & 6.664 & 7.044 & 3.900 & 0.027 \\
P4 & 9.686 & 17.729 & 8.664 & 4.306 & 0.031 \\
P5 & 8.861 & 8.611 & 8.099 & 5.246 & 0.042 \\
P6 & 7.854 & 6.286 & 7.184 & 3.442 & 0.033 \\
P7 & 8.564 & 5.499 & 7.918 & 3.540 & 0.033 \\
P8 & 8.018 & 5.955 & 7.340 & 3.742 & 0.031 \\
P9 & 9.720 & 17.867 & 8.703 & 4.305 & 0.028 \\ \bottomrule
\end{tabular}%
}
\caption{T-test statistics for average fixation duration time per word for Positive labels (Cognates) and Negative labels (False Friends) for participants P1-P9.}
\label{tab:t-test}
\end{table}
\subsection{Gaze Behaviour Data Analysis}

The accuracy of similarity annotation by participants lies between 98\% to 99.5\% for individual annotators. Out of the 1800 annotations (9 annotators over 200 word-pairs), only 40 were predicted incorrectly.  Annotation errors may be attributed to: (a) lack of patience/attention while reading, (b) issues related to word-pair comprehension, and (c) confusion/indecisiveness caused due to lack of contextual clues. In our analysis, we do not discard the data obtained from these incorrect annotations. 

We observe distinct eye-movement patterns for cognate pairs in terms of fixation duration of the human readers. Our analysis shows that fixation duration normalized over word count is relatively larger for cognate pairs. For cognate pairs, we observe that average fixation duration amongst all participants is 1.3 times more than that of false friend pairs. To test the statistical significance, we conduct a two-tailed t-test (assuming unequal variance) to compare the average fixation duration per word for cognate and false friend pairs. The hypothesized
mean difference is set to 0, and the error tolerance limit ($\alpha$) is set to 0.05. The t-test analysis, presented in Table \ref{tab:t-test}, shows that for all participants, a statistically significant difference exists between the average fixation duration per word for cognate pairs vs false friend pairs. 

We believe this difference in average fixation duration is because the bilingual speakers who participated in the experiment can clearly distinguish between cognates and false friends and decide quickly in either case. The duration for which they fixate on either of the cases differs significantly for each participant. As per our observation, the participants take more time over cognate pairs to ensure similarity in meaning. Given their knowledge of both the languages and contextual clues, they were 'quickly' able to decide these word pairs did not mean the same. It should be noted that they were unaware of the 'cognate' or 'false friend' distinction concerning the experiment. They were simply asked to note the meaning of both the words given context, and annotate accordingly, as described in the paper.

\subsection{Cross-lingual Word Embeddings}

For this task, we use the cross-lingual word embedding models released by~\citet{kumar-etal-2020-passage}. The Hindi-Marathi cross-lingual models released with this paper are based on both MUSE~\cite{conneau2017word} and XLM\footnote{Word representations extracted from the last layer of the contextual XLM model.}~\cite{lample2019cross}. Additionally, we build the cross-lingual word embeddings model for Hindi-Marathi using VecMap~\cite{artetxe-etal-2017-learning}. The model uses monolingual corpora released by~\citet{kunchukuttan2020ai4bharat} and a bilingual dictionary\footnote{\href{http://www.cfilt.iitb.ac.in/Downloads.html}{Bilingual Lexicon}} required for the supervised method by~\citet{artetxe-etal-2017-learning}. These three cross-lingual models provide us with three feature sets for the task of cognate detection. 

\section{Feature Sets for Cognate Detection}

In this section, we discuss the various features used for the task of cognate detection. It is to be noted that \textit{false friends are spelt similarly across languages} but mean differently. Using false friends as data points with negative labels restricts us to the use of semantic similarity based features, as orthographic or phonetic similarity-based measures would fail to detect sufficient distinction between them. Hence, we use the features proposed by~\citet{rama2016siamese} and~\citet{kanojia2019cognate} as baseline features for a comparative evaluation.

\subsection{Phonetic Features}

The IndicNLP Library provides phonetic features based vector for each character in various Indian language scripts. We utilize this library to compute a feature vector for each word by computing an average over character vectors. We compute vectors for both words in the candidate cognate pairs ($PV_S$ and $PV_T$) and also compute contextual vectors ($PCV_S$ and $PCV_T$) by averaging the vectors for all the contextual clues, generating a total of four vectors. We use these vectors as features for computing the baseline scores using the Siamese Convolutional Neural Network architecture proposed by~\citet{rama2016siamese}.

\subsection{Weighted Lexical Similarity (WLS)}

The Normalized Edit Distance (NED) approach computes the edit distance~\cite{nerbonne1997measuring} for all word pairs in our dataset. A combination of NED with q-gram distance~\cite{shannon1948mathematical} for a better similarity score. The q-grams (`n-grams') are simply substrings of length q. This distance measure has been applied previously for various spelling correction approaches~\cite{owolabi1988fast,kohonen1978very}.~\citet{kanojia2019utilizing} proposed Weighted Lexical Similarity (WLS) and we use it with the character-based Recurrent Neural Network architecture proposed by them to compute another set of baseline scores.

\subsection{Cross-lingual Vectors \& Similarity}

As discussed above, we use the pre-trained cross-lingual embedding models for generating feature vectors for MUSE and XLM based approaches. These models are generated by aligning two disjoint monolingual vector spaces through linear transformations, using a small bilingual dictionary for supervision~\cite{doval2018improving,artetxe-etal-2017-learning}. Additionally, the cross-lingual embeddings model trained using~\citet{artetxe-etal-2017-learning}'s approach provides us with the third set of feature vectors. 

We use these models to obtain vectors for word-pairs ($WV_S$ and $WV_T$) and averaged context vectors ($CV_S$ and $CV_T$) from the contextual clues, to create three different feature sets. We obtain vectors for each candidate pair and their context using all the three cross-lingual methodologies. The use of cross-lingual models has been proposed for differentiating between cognates and false friends by~\citet{merlo-andueza-rodriguez-2019-cross}, but evaluation with the cognate detection task had not been performed. We perform this evaluation on our datasets (D1, D2 and D1+D2) using various classification methods and discuss them later.

\begin{table*}[]
\centering
\resizebox{1\textwidth}{!}{%
\begin{tabular}{cc}
\hline
\textbf{Gaze Feature} & \textbf{Description} \\ \hline
Average Fixation Duration & The average of all fixation duration across all interest areas present on the screen. \\
Average Saccade Amplitude & Saccade amplitude is the amplitude of going back and forth measured in terms of duration. \\
Fixation Count & Counting the number of times user's eyes are fixated on the screen. \\
Fixation Duration Max & Maximum time for a single fixation on any Interest Area. \\
Fixation Duration Min & Minimum time for a single fixation on any Interest Area. \\
IA Count & Interest Area Count (no. of IAs on the screen) \\
Run Count & Consecutive counts for same Interest Area are ignored in Run Count \\
Saccade Count & Total counts of Saccades \\ \hline
\end{tabular}%
}
\caption{Gaze Features used for the task of Cognate Detection}
\vspace{-0.5cm}
\label{tab:gazefeat}
\end{table*}

\subsection{Cognitive Features from Gaze Data}

Gaze behaviour of participants, characterized by fixations, forward saccades, skips and regressions, can be used as features for NLP tasks~\cite{mishra-etal-2018-author}. Since these gaze features relate to the cognitive process in reading~\cite{altmann1994regression}, we consider these as features in our model. The current gaze data extraction adds up all the interest areas on the screen (word + context). The software used to analyze the gaze data, currently, provides us with collated results for all the gaze-based features. 

From the gaze behaviour data collected, we extract a total of 18 features for each of the 1800 data points. Using supervised feature selection approach, we are able to select eight best features via grid search using Logistic Regression. We use the SelectKBest implementation along with hyperparameter tuning via GridSearchCV, present in the sklearn~\cite{pedregosa2011scikit} library. Here onwards, we refer to these eight features when discussing cognitive or gaze-based features in this paper. These eight best features, along with their description, are listed in Table \ref{tab:gazefeat}. 

\subsection{Gaze Feature Prediction}
\label{subsec:gazepredict}
Collection of gaze data for a large number of samples can be a costly task. We propose a neural model for cognitive features prediction. Our neural model is a feed-forward neural network to perform a regression task and predict gaze features. We collect gaze data for only 200 word-pairs (D2) with the help of 9 annotators which provides us with a total of 1800 data points for training and validation. As reported in Table \ref{tab:results1}, the initial results on D1 using different cross-lingual embeddings show that XLM based contextual features perform the best amongst all the cross-lingual models. 

\begin{figure}[ht]
    \centering
    \includegraphics[width=1\columnwidth]{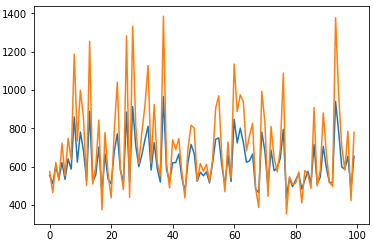}
    \caption{Predicted feature values ( blue ) vs. Gold feature values ( orange ) for the average fixation duration feature, on 100 samples.}
    \label{fig:saccade}
    \vspace{-0.4cm}
\end{figure}

As an input to the network, we provide the feature vectors from the XLM model. This network's output is the predicted gaze features for the D1 dataset, using the gaze features as gold predictions from the D2 dataset. This network contains three linear hidden layers with 128, 64 and 32 dimensions. After each layer, we use the sigmoid activation function and dropout after each sigmoid with a dropout value of 0.2. We use 0.1 as the learning rate and use the Mean Squared Error (MSE) loss function. A graph comparing the values for the predictions vs the actual values for the average saccade amplitude for 100 samples, can be seen in Figure \ref{fig:saccade}.

\section{The Cognate Detection Task}

We employ both classical machine learning-based models and a simple feed-forward neural network. To compare our work with the previously proposed approaches, we replicate the best-reported systems from \newcite{rama2016siamese} \textit{i.e.,} Siamese Convolutional Neural Network with phonetic vectors as features and also replicate \newcite{kanojia2019utilizing}'s approach which uses a Recurrent Neural Network architecture with a weighted lexical similarity (WLS) as the feature set. The input to our classifiers is the feature sets described above for each candidate pair. 

Among the classical machine learning models, we use Support Vector Machines (SVM) and Logistic Regression (LR). We experiment with the use of both linear SVMs and kernel SVMs (Gaussian and Polynomial). We perform a grid-search to find the best hyper-parameter value for $C$ over the range of 0.01 to 1000. We deploy the Feed Forward Neural Network (FFNN) with one hidden layer. We perform cross-validation with different activation function settings (tanh, hardtanh, sigmoid and relu) and the hidden layer dimension in the network (30, 50, 100, and 150). We use binary cross-entropy as the optimization algorithm. Finally, we choose the hyper-parameter configuration with the best validation accuracy. We train the model with the selected configuration with an initial learning rate of 0.4, and we halve the learning rate when the error on the validation split increases. We stop the training once the learning rate falls below $0.001$. \textit{We perform 5-fold stratified cross-validation, which divides the data into train and test folds, randomly.} 

Initially, we perform our experiments with the feature sets from three different cross-lingual embeddings (MUSE, XLM, and VecMap) for the dataset D1, then with the smaller dataset D2 and later on the combined dataset D1+D2. We, then, perform the same task for the smaller dataset D2 by combining cognitive features with individual cross-lingual feature sets. We also observe the performance of standalone gaze features for the D2 dataset. Finally, we evaluate the predicted gaze features on the combined dataset by combining them with cross-lingual features and a standalone feature set, using the feed-forward neural network. We report the results of the cognate detection task in the next section and discuss them in detail.

\begin{table*}[ht]
\centering
\resizebox{1.0\textwidth}{!}{%
\begin{tabular}{@{}ccccccccccccc@{}}
\toprule
 & P & R & F & P & R & F & P & R & F & P & R & F \\ \midrule
Feature Set $\rightarrow$ & \multicolumn{3}{c}{Phonetic} & \multicolumn{3}{c}{WLS} &  &  &  &  &  &  \\ \midrule
Rama et. al., 2016 (D1+D2) & 0.71 & 0.69 & 0.70 & - & - & - &  &  &  &  &  &  \\
Kanojia et. al., 2019 (D1+D2) & - & - & - & 0.76 & 0.72 & 0.74 &  &  &  &  &  &  \\ \midrule
Feature Set $\rightarrow$ & \multicolumn{3}{c}{XLM} & \multicolumn{3}{c}{MUSE} & \multicolumn{3}{c}{VecMap} &  &  &  \\ \midrule
Linear SVM (D1+D2) & 0.83 & 0.71 & 0.77 & 0.72 & 0.68 & 0.70 & 0.70 & 0.65 & 0.67 &  &  &  \\
LogisticRegression (D1+D2) & 0.85 & 0.74 & 0.79 & 0.80 & 0.71 & 0.75 & 0.70 & 0.66 & 0.68 &  &  &  \\
FFNN (D1+D2) & 0.82 & 0.84 & \textbf{0.83} & 0.83 & 0.79 & 0.81 & 0.75 & 0.76 & 0.75 &  &  &  \\ \midrule
Feature Set $\rightarrow$ & \multicolumn{3}{c}{XLM+Gaze} & \multicolumn{3}{c}{MUSE+Gaze} & \multicolumn{3}{c}{VecMap+Gaze} & \multicolumn{3}{c}{Gaze} \\ \midrule
Linear SVM (D2) & 0.81 & 0.69 & 0.75 & 0.72 & 0.73 & 0.72 & 0.70 & 0.75 & 0.72 & 0.77 & 0.76 & 0.76 \\
LogisticRegression (D2) & 0.84 & 0.75 & 0.79 & 0.76 & 0.72 & 0.74 & 0.81 & 0.71 & 0.76 & 0.80 & 0.75 & 0.77 \\
FFNN (D2) & 0.83 & 0.85 & \textbf{0.84} & 0.83 & 0.78 & 0.80 & 0.86 & 0.83 & 0.84 & 0.81 & 0.71 & 0.76 \\ \midrule
\multicolumn{13}{c}{\textbf{Predicted Gaze Features On D1 (11652 samples) and Collected Gaze Features on D2 (200 samples)}} \\ \midrule
Feature Set $\rightarrow$ & \multicolumn{3}{c}{XLM+Gaze} & \multicolumn{3}{c}{MUSE+Gaze} & \multicolumn{3}{c}{VecMap+Gaze} & \multicolumn{3}{c}{Gaze} \\ \midrule
FFNN (D1+D2) & 0.84 & 0.88 & \textbf{0.86} & 0.85 & 0.78 & 0.81 & 0.83 & 0.85 & 0.84 & 0.77 & 0.76 & 0.76 \\ \midrule
FFNN (D1) [Only Predicted Gaze] & 0.83 & 0.84 & 0.83 & 0.82 & 0.79 & 0.80 & 0.80 & 0.86 & 0.83 & 0.76 & 0.77 & 0.76 \\ \bottomrule
\end{tabular}%
}
\caption{Classification results in terms of weighted Precision (P), Recall (R), and F-scores (F) using 5-fold cross-validation using different feature sets as described above.}
\label{tab:results1}
\end{table*}

\begin{table*}[h]
\centering
\resizebox{0.9\textwidth}{!}{%
\begin{tabular}{@{}cccccccccc@{}}
\toprule
 & \multicolumn{3}{c}{Phonetic} & \multicolumn{3}{c}{WLS} &  &  &  \\ \midrule
 & P & R & F & P & R & F & P & R & F \\ \midrule
Rama et. al., 2016 (D1) & 0.70 & 0.68 & 0.69 & - & - & - &  &  &  \\
Kanojia et. al., 2019 (D1) & - & - & - & 0.74 & 0.70 & 0.72 &  &  &  \\ \midrule
Rama et. al., 2016 (D2) & 0.64 & 0.57 & 0.60 & - & - & - &  &  &  \\
Kanojia et. al., 2019 (D2) & - & - & - & 0.61 & 0.66 & 0.63 &  &  &  \\ \midrule
 & \multicolumn{3}{c}{XLM} & \multicolumn{3}{c}{MUSE} & \multicolumn{3}{c}{VecMap} \\ \midrule
Linear SVM (D1) & 0.81 & 0.71 & 0.76 & 0.70 & 0.68 & 0.69 & 0.70 & 0.65 & 0.67 \\
LogisticRegression (D1) & 0.80 & 0.75 & 0.77 & 0.72 & 0.74 & 0.73 & 0.70 & 0.73 & 0.71 \\
FFNN (D1) & 0.80 & 0.84 & 0.82 & 0.81 & 0.76 & 0.78 & 0.77 & 0.76 & 0.76 \\ \midrule
Linear SVM (D2) & 0.72 & 0.65 & 0.68 & 0.65 & 0.60 & 0.62 & 0.62 & 0.57 & 0.59 \\
LogisticRegression (D2) & 0.78 & 0.69 & 0.73 & 0.67 & 0.67 & 0.67 & 0.63 & 0.61 & 0.62 \\
FFNN (D2) & 0.79 & 0.81 & 0.80 & 0.76 & 0.71 & 0.73 & 0.74 & 0.71 & 0.72 \\ \bottomrule
\end{tabular}%
}
\caption{Additional results in terms of weighted Precision (P), Recall (R), and F-scores (F) using 5-fold cross-validation using different feature sets as described above. These are additional results on the individual datasets D1 and D2 for which the combined results (D1 and D2) are already shown in Table \ref{tab:results1} for a fair comparison.}
\label{tab:results2}
\end{table*}

\section{Results and Dicussion}

We report the results of the cognate detection task in Table \ref{tab:results1}. We use the original implementations of \citet{rama2016siamese} and \citet{kanojia2019utilizing} on the combined (D1+D2) dataset. In our initial evaluation on the D1 dataset, cross-lingual model-based features (XLM, MUSE, and VecMap) can be seen to outperform the baseline systems which use phonetic and orthographic features. Using the XLM-based features, we observe an improvement of 9\% over the stronger baseline~\cite{kanojia2019utilizing} and 13\% over the system by \newcite{rama2016siamese}. It can be seen that MUSE and VecMap based features also perform better on the combined dataset. In terms of both precision and recall, cross-lingual features are shown to outperform the baseline systems. The cross-lingual approach, with representations from VecMap-based model, fails to perform as well as MUSE and XLM-based models. The contextual XLM model achieves the best scores in almost all the settings. We believe its performance can be attributed to the linguistic closeness of the language pair and context from the contextual clues provided. For example, the false friend pair ``\textit{kaccHa}'' (meaning inexperienced) - ``\textit{kaccHa}'' [raw (food)] is classified correctly by XLM but incorrectly by both the baseline models, and VecMap-based classification.  This signifies that fine-grained semantic difference between such false friend pairs can be captured via cognitive features. We report the additional results on individual datasets (D1 and D2), for all the baseline and cross-lingual approaches, in Table \ref{tab:results2}.

For all the classifiers, the gaze features are averaged across participants and augmented with cross-lingual features. The gaze fixation duration collects the total time spent, as fixations, on each interest area including the context clues. We were hopeful that the participants would focus only on important contextual clues and not the stop words with our experiment design. However, the sample points are not enough to concretely discuss this aspect of our study. These results are reported for all the classifiers with D2 dataset. Our feature combinations outperform the baselines with an F-score improvement of 10\% points over the stronger baseline (WLS). We also report the precision, recall and F-score values when only gaze features are used to predict the labels for our candidate pairs. We observe that standalone gaze features are not as effective as when combined with cross-lingual feature vectors. When gaze features are predicted using the methodology described in Section \ref{subsec:gazepredict}, the model performance for FFNN on D1 remains the same with XLM+Gaze, decreases slightly for MUSE+Gaze and significantly improves for VecMap+Gaze. We observe that predicted gaze features do not significantly drop the performance, and hence, add the collected gaze data samples (D2) to D1. 

On the combined dataset with collected gaze features (on D2) and predicted gaze features (on D1), we report our best system [FFNN (D1+D2)] which shown an improvement of 12\% over the stronger baseline (WLS), and 16\% over the weaker baseline (Phonetic). This system also outperforms the best reported cross-lingual features-based approach by 3\%, as shown in Table \ref{tab:results1}.

For example, the cognate pair ``\textit{utPaNa}'' (Hindi) - ``\textit{utpAaDit}'' (Marathi) (both meaning manufactured) is classified correctly by this system, but incorrectly by both the baselines, and all the cross-lingual systems. Furthermore, we also show that gaze features can be predicted based on a small sample data, and improved performance can be attained with the help of cross-lingual features, reported with our work.

\section{Conclusion and Future Work}

In this paper, we harness cross-lingual embeddings and gaze-based features to improve the task of cognate detection for the Indian language pair of Hindi-Marathi. We create a novel framework that derives insights from human cognition, that manifests over eye movement patterns. We hypothesize that by augmenting cross-lingual features with features obtained from the gaze data, the task of cognate detection can be improved. We use a linked knowledge graph (IndoWordnet) to augment a publicly released cognate dataset with contextual clues. We collect the gaze behaviour data from nine participants over 200 samples and perform the task of cognate detection for both our datasets (with gaze data and without gaze data). Then, we use a neural network to predict gaze features for unseen samples and perform the task of cognate detection to show improved performance, despite a small sample of collected gaze data.

We reproduce the previously proposed baseline approaches and perform experiments using additional features obtained via cross-lingual models for a comparative evaluation. The previously proposed approaches~\cite{rama2016siamese,kanojia2019utilizing} for this task are shown to be outperformed by cross-lingual features and the combination of these features with the obtained gaze data. Our experiments use three different approaches to generate feature representations for the cognate detection task, and all of them show improvements over previously proposed approaches. We observe consistent improvements in terms of precision, recall and F-scores. Over the stronger baseline, our best system shows an improvement of 12\% points and 16\% points over the weaker baseline. This system also outperforms the cross-lingual features based approaches by 3\%, over the combined dataset. We release this augmented dataset, along with our code and cross-lingual models for further research.

In future, we aim to add more language pairs and leverage contextual information from knowledge graphs using sequence-based neural models. We also aim to collect gaze data and then model the gaze predictions in a multi-task setting. We plan to investigate other multilingual contextual embeddings' performance for this task (\textit{e.g.,} M-BERT, IndicBERT, MuRIL). We also plan to look for a method to differentiate between different interest areas and see if a markup facility is present in the software used to analyze the gaze data. We also aim to investigate the task of cognate detection for the Indo-European language family, in the near future.

\bibliographystyle{acl_natbib}
\bibliography{anthology,eacl2021}

\end{document}